\documentclass[11pt]{article}
\usepackage{graphicx}	
	\newcommand{\blind}{0}
	
    \makeatletter
    \renewcommand\section{\@startsection {section}{1}{\z@}%
                                       {-3.5ex \@plus -1ex \@minus -.2ex}%
                                       {2.3ex \@plus.2ex}%
                                       {\normalfont\fontfamily{phv}\fontsize{16}{19}\bfseries}}
    \renewcommand\subsection{\@startsection{subsection}{2}{\z@}%
                                         {-3.25ex\@plus -1ex \@minus -.2ex}%
                                         {1.5ex \@plus .2ex}%
                                         {\normalfont\fontfamily{phv}\fontsize{14}{17}\bfseries}}
    \renewcommand\subsubsection{\@startsection{subsubsection}{3}{\z@}%
                                        {-3.25ex\@plus -1ex \@minus -.2ex}%
                                         {1.5ex \@plus .2ex}%
                                         {\normalfont\normalsize\fontfamily{phv}\fontsize{14}{17}\selectfont}}
    \makeatother
	
	\usepackage{amsmath}
	\usepackage{graphicx}
	\usepackage{enumerate}
	\usepackage{natbib} 
	\usepackage{url} 
	
\begin{document}
		
		\def\spacingset#1{\renewcommand{\baselinestretch}%
			{#1}\small\normalsize} \spacingset{1}
		
		\if0\blind
		{
			\title{\bf \emph{RoBERTa-wwm-ext Fine-Tuning for Chinese Text Classification} }
			\author{Zhuo Xu \\
			The Ohio State University - Columbus \\
             xu.3614@osu.edu }
			\date{}
			\maketitle
		} \fi
		
		\if1\blind
		{

            \title{\bf \emph{IISE Transactions} \LaTeX \ Template}
			\author{Author information is purposely removed for double-blind review}
			
\bigskip
			\bigskip
			\bigskip
			\begin{center}
				{\LARGE\bf \emph{IISE Transactions} \LaTeX \ Template}
			\end{center}
			\medskip
		} \fi
		\bigskip
		
	\begin{abstract}
Bidirectional Encoder Representations from Transformers (BERT) have shown to be a promising way to dramatically improve the performance across various Natural Language Processing tasks \cite{devlin}. Meanwhile, progress made over the past few years by various Neural Network has also proved the effectiveness of Neural Network in the field of Natural Language Processing. In this project, RoBERTa-wwm-ext\cite{cui} pre-train language model was adopted and fine-tuned for Chinese text classification. The models were able to classify Chinese texts into two categories, containing descriptions of legal behavior and descriptions of illegal behavior. Four different models are also proposed in the paper. Those models will use RoBERTa-wwm-ext as their embedding layer and feed the embedding into different neural networks. The motivation behind proposing these models is straightforward. By introducing complex output layer architecture, the overall performance of the models could be improved. All the models were trained on a dataset derived from Chinese public court records, and the performance of different models were compared. The experiment shows that the performance of proposed models failed to beat the original RoBERTa-wwm-ext model in terms of accuracy and training efficiency.
	\end{abstract}
			
	\noindent%

	\spacingset{1.5} 

\section{Introduction} \label{s:intro}
With the increasing use of social media in recent years, the number of cybercrime cases are growing. Nowaday, there are multitudinous social platforms with large number of participants, making the supervision of cybercrime very difficult. In China, the traditional way to monitor cybercrime is to hire a group of people and manually identify these illegal behaviors online. However, with more than 900,000,000 citizens connected to the internet, this type of monitoring system would require big budgets and huge staff. Therefore, a model that can automatically identify the content containing potential illegal behaviors is much-needed.

The models use RoBERTa-wwm-ext instead of the original BERT because RoBERTa-wwm-ext will bring more performance gain. RoBERTa-wwm-ext model was believed to have two major optimizations compared to the original BERT will be beneficial to the classification task. The first one is the use of RoBERTa pre-trained model. \cite{roberta} believe that the original BERT training is inadequate. They conducted a complex study, including careful evaluation of the effects of hyperparameters and size of training set on the BERT pre-training model, and optimized the model based on their finding. The second one is pre-trained the model with whole word masking for Chinese. Since the models will be applied to classify Chinese text, whole-word Masking will allow the models to incorporate Chinese language scenarios and apply them to knowledge graphs to provide more information about Chinese semantics.

The contributions of this paper are as follow:
\begin{itemize}
\item Four different models were proposed to fine-tune the pre-trained RoBERTa-wwm-ext model to classify whether a piece of text describes illegal behavior. 
\item After researching the fine-tuning method for BERT, pre-process long text was adopted because the models could yield better performance. 
\item Training hyper-parameters were fine-tuned to achieve a balance of training time and performance.   
\end{itemize}

\section{Data Processing} \label{s:sec2}
The raw date of training set was obtained from public court record. The record is available at the following link:
\begin{quote}
\url{https://susong.tianyancha.com/}
\end{quote}
The data processing steps are as follow:
First, all samples were extracted from raw data. Then sentences containing illegal behaviors were manually labeled with 1. Those sentences will become the positive example dataset. The rest of sentences were labeled with 0 and used as negative examples. Positive and negative samples were then randomly mixed.
In total, the dataset contained 6755 samples. Datasets were then be randomly split into test set and training set, where test set account for 20\% and the rest of data will become a traning set. The training set was further partitioned into a new training set and a validation set. The new training set will contain 64\% of the whole dataset, and validation set will contain 16\% of the whole dataset.


\section{RoBERTa-wwm-ext model Fine-Tuning for Chinese Text Classification}
\label{s:methods}
\subsection{\emph{Methodology}} \label{s:methods.1}
For the baseline model, the pre-trained RoBerta-wwm-ext was fine-tuned using the data set mentioned above. The output from the transformers was fed into a fully connected classification layer and then the model will then output whether the classification result is right or wrong.

In terms of the four models proposed, the output from the transformer were feed into four different types of Neural Network. Those Neural Network includes a Convolutional Neural Network \cite{DBLP}, a Recurrent Neural Network \cite{LiuQH}, a Recurrent Convolutional Neural Network \cite{rcnn}, and a Deep Pyramid Convolutional Neural Network\cite{johnson-zhang-2017-deep}. All the models were implemented using Pytorch.
\subsection{\emph{Model Fine-Tuning with Convolutional Neural Network as classification layer}} \label{s:methods.2}
In these models, the incoming text data will be padded into the same length, and then it will be fed into pre-trained RoBerta-wwm-ext encoder to get the words embedding. None of the layers of the pre-trained RoBerta-wwm-ext model were frozen in the training process. The embedding will then be feed into a Convolutional Neural Network. The convolution layer has three different sizes of convolutions kernels which are 2, 3 and 4. For each kernel size, there are 100 convolution kernels. This allows the model to extract 2-gram, 3-gram, 4-gram feature from embedding. Those features will be feed into a max-pooling layer to filter out unimportant information. After that, those three feature maps will be concatenated and fed into a linear fully connected layer and output the classification result. The dropout rate set for the Convolutional Neural Network is 0.1.\cite{DBLP} 

\subsection{\emph{Model Fine-Tuning with Recurrent Neural Network as classification layer}} \label{s:methods.3}
In this model, the incoming text data will also be padded into a certain length, and then it will be feed into pre-trained RoBerta-wwm-ext encoder to get the words embedding. None of the layers of the pre-trained RoBerta-wwm-ext model were frozen in the training process as well. The embedding would be fed into a Recurrent Neural Network which consist of two layers of Bidirectional long short-term memory (BiLSTM) with a hidden size of 768. After that, the last hidden state from BiLSTM layer would be fed into a linear fully connected layer and output the classification result. The dropout rate set for the Recurrent Neural Network is 0.1.\cite{LiuQH} 
\subsection{\emph{Model Fine-Tuning with Recurrent Convolutional Neural Network as classification layer}} \label{s:methods.4}
In this model, the incoming text data will also be padded into length of certain, and then it will be feed into a pre-trained RoBerta-wwm-ext encoder to get the words embedding. The RoBerta-wwm-ext layers in this model would not be frozen either. The embedding produced by the encoder would be fed into a Recurrent Neural Network which consists of two layers of Bidirectional long short-term memory (BiLSTM) with a hidden size of 768.  After that, the output from BiLSTM layer and the embedding produced by RoBerta-wwm-ext were concatenated and fed into a ReLU activation layer. The output of activation layer would be fed into a max-pooling layer to filter out the noise. The output from max-pooling layer will then be fed into a linear fully connected layer and output the classification result. The dropout rate for the Recurrent Convolutional Neural Network was set to be 0.1.\cite{rcnn}
\subsection{\emph{Model Fine-Tuning with Deep Pyramid Convolutional Neural Network as classification layer}} \label{s:methods.5}
In this model, the incoming text data will be padded into a certain length, and then it will be feed into pre-trained RoBerta-wwm-ext encoder to get the words embedding. All layers of the models, including the RoBerta-wwm-ext encoder, were trained. The embedding will then be feed into a Deep Pyramid Convolutional Neural Network. The embedding will first go through a process called region embedding. In this process, the embedding would be fed into a convolution layer that has 250 kernels, and each kernel's size is 3. In order to keep the length of sequence unchanged, the output will be padded and fed into a ReLU activation layer. After that, it would be fed into a convolution layer. The output of this convolution layer will go through being padded, feeding into the ReLU activation layer, and convolution layer again. After the process is completed, the convolution layer's output will then be fed into following process, and this process would be repeated several times. The process has several steps: 

\begin{itemize}
\item The input will go through a Max\_pooling layer with a kernel with size of 3 and stride of 2, which essentially halves the length of input.
\item Then the output will be subject to two convolution layer's operation with 250 kernels of size 3 in each layer.
\item A residual connection of outputs from previous two steps was conducted.
\end{itemize}
\includegraphics[scale=0.6]{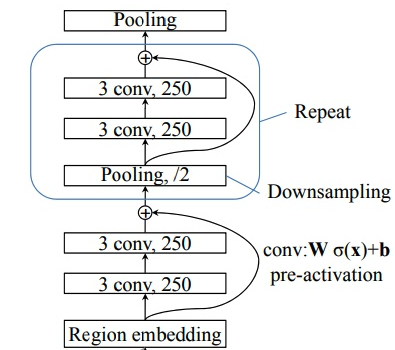}

The output from the process will then be fed into a linear fully connected layer and output the classification result. The dropout rate for the Deep Pyramid Convolutional Neural Network was set as 0.1.\cite{johnson-zhang-2017-deep} 
\subsection{\emph{Model Fine-Tuning with fully connected classification layer}} \label{s:methods.6}
This model will act as the baseline model of the experiment. The incoming text data will be padded into a certain length, and then it will be feed into pre-trained RoBerta-wwm-ext encoder. The output of the encoder will be fed into a linear fully connected layer and output the classification result.
\section{Experiment}\label{s:methods.7}
\subsection{\emph{Experiment Enviorment}} \label{s:methods.6}	
All of the model training was conducted on a node of OSC Owens Cluster. Each node has 28 CPU cores from Intel Xeon E5-2680 v4, 128 GB of memory, and one NVIDIA Tesla P100 GPU with 16GB memory. CUDA10.2.89 was also enabled to accelerate training process. 

\begin{table}[!htbp]
\centering
\begin{tabular}{|c|c|c|}
\hline \textbf{Training time} & \textbf{Batch Size} & \textbf{Model} \\ \hline
 00:04:02 & 64 & Baseline \\
 00:04:15 & 64 & CNN\\
 00:05:42 & 64 & RNN\\
 00:03:46 & 64 & DPCNN\\
 00:04:10 & 64 & RCNN\\
\hline
\end{tabular}
\caption{\label{time} Training time for running 10 epochs. (Model types are denoted by the classification layer it adopted)}
\end{table}

\subsection{\emph{Training}} \label{s:methods.6}
Table~\ref{time} shows the training time for training these proposed models with batch size 64. It could be observed that the training time for these models did not vary by a notable amount, except for the case of Recurrent Neural Network as the classifier layer of RoBERTa-wwm-ext. In which case, the model took a little bit longer to train than other models proposed.
Table~\ref{baseline} \ref{cnn} \ref{dpcnn} \ref{rcnn} \ref{rnn} shows the training time of different batch size for different model. It shows that train the model with smaller batch size did not improve accuracy very much. In this case, larger batch size is preferred because similar performance could be achieved in a much shorter training time. Due to memory constraint of the GPU, experiments with larger batch size were not completed, as CUDA out of memory error was encountered when trying to train the model.

\begin{table}[!htbp]
\centering
\begin{tabular}{|c|c|c|}
\hline \textbf{Training time} & \textbf{Batch Size} & \textbf{Val Acc} \\ \hline
 00:04:02 & 64 & 95.31\% \\
 00:12:58 & 16 & 96.58\% \\
\hline
\end{tabular}
\caption{\label{baseline} Training time for baseline model running 10 epochs.}
\end{table}

\begin{table}[!htbp]
\centering
\begin{tabular}{|c|c|c|}
\hline \textbf{Training time} & \textbf{Batch Size} & \textbf{Val Acc} \\ \hline
 00:04:15 & 64 & 94.36\% \\
 00:13:12 & 16 & 94.62\% \\
\hline
\end{tabular}
\caption{\label{cnn} Training time for RoBerta-wwm-ext CNN model running 10 epochs.}
\end{table}

\begin{table}[!htbp]
\centering
\begin{tabular}{|c|c|c|}
\hline \textbf{Training time} & \textbf{Batch Size} & \textbf{Val Acc} \\ \hline
 00:04:10 & 64 & 94.14\% \\
 00:12:52 & 16 & 94.68\% \\
\hline
\end{tabular}
\caption{\label{rcnn} Training time for RoBerta-wwm-ext RCNN model running 10 epochs.}
\end{table}

\begin{table}[!htbp]
\centering
\begin{tabular}{|c|c|c|}
\hline \textbf{Training time} & \textbf{Batch Size} & \textbf{Val Acc} \\ \hline
 00:03:46 & 64 & 92.38\% \\
 00:12:43 & 16 & 93.36\% \\
\hline
\end{tabular}
\caption{\label{dpcnn} Training time for RoBerta-wwm-ext DPCNN model running 10 epochs.}
\end{table}

\begin{table}[!htbp]
\centering
\begin{tabular}{|c|c|c|}
\hline \textbf{Training time} & \textbf{Batch Size} & \textbf{Val Acc} \\ \hline
 00:05:42 & 64 & 95.03\% \\
 00:13:19 & 16 & 94.63\% \\
\hline
\end{tabular}
\caption{\label{rnn} Training time for RoBerta-wwm-ext RNN model running 10 epochs.}
\end{table}

\subsection{\emph{Performance}} \label{s:methods.6}
Performance wise, Table~\ref{baseline} \ref{cnn} \ref{dpcnn} \ref{rcnn} \ref{rnn} shows that the models proposed weren't able to excels the baseline RoBERTa-wwm-ext model. In terms of accuracy, none of the models was able to present superiority over the RoBERTa-wwm-ext model used as base line, except when Recurrent Neural Network as the classifier layer of RoBERTa-wwm-ext the validation accuracy came very close to the performance of the baseline model. In terms of training time, the proposed models also failed to complete training in a shorter time.
\section{Conclusion}\label{s:conclusion}
In this paper, four different models to fine-tune RoBERTa-wwm-ext were proposed. During the experiments, the performance of the RoBERTa-wwm-ext encoder was found to be very strong. In the proposed models, RoBERTa-wwm-ext was adopted as embedding layer, and the embedding would be fed into four different types of neural networks. These neural networks include Convolutional Neural Network, Deep Pyramid Convolutional Neural Network, Recurrent Convolutional Neural Network, and Recurrent Neural Network. Those models with neural network as output layers failed to outperform the baseline model in accuracy and training efficiency.

\bibliographystyle{chicago}
\spacingset{1}
\bibliography{IISE-Trans}
	
\end{document}